%% file: ChiMDQA_en.tex
\documentclass[runningheads]{llncs}
\usepackage[T1]{fontenc} 
\usepackage{graphicx}
\usepackage{wrapfig}
\usepackage{lipsum} 
\usepackage{wrapfig}
\usepackage{float} 
\usepackage{booktabs}  
\usepackage{amssymb}   
\usepackage{graphicx}  
\usepackage[numbers]{natbib}  
\input{math_commands.tex}

\usepackage{hyperref}
\usepackage{url}
\usepackage{multicol}
\usepackage{aliascnt}
\usepackage{adjustbox}
\usepackage{tcolorbox}
\usepackage{booktabs}
\usepackage{multirow}
\usepackage{tabularx} 
\usepackage{enumitem}
\usepackage{makecell} 
\usepackage{array}    
\definecolor{uclablue}{rgb}{0.15, 0.45, 0.68}
\hypersetup{
    breaklinks,
    citecolor=uclablue,
    colorlinks=true,
}
\usepackage{booktabs} 
\usepackage[table]{xcolor} 

\usepackage[margin=1in]{geometry}  
\usepackage[fontsize=10pt]{fontsize}

\usepackage{diagbox} 
\usepackage{amsmath} 
\usepackage{caption} 

\begin{document}
\title{ChiMDQA: Towards Comprehensive Chinese Document QA \\with Fine-grained Evaluation}
\titlerunning{ChiMDQA: Towards Comprehensive Chinese Document QA with Fine-grained Evaluation}
\author{
    Jing Gao\inst{1,4} \and
    Shutiao Luo\inst{2,4} \and
    Yumeng Liu\inst{3,4} \and
    Yuanming Li\inst{4}\thanks{Corresponding author.} \and
    Hongji Zeng\inst{4}
}

\institute{
    Beijing Jiaotong University, Beijing, China \and
    Beijing University of Posts and Telecommunications, Beijing, China \and
    Beijing University of Technology, Beijing, China \and
    Foxit Software Co. Ltd, Fuzhou, China
    \\
    \email{jennygao@bjtu.edu.cn, luoshutiao@bupt.edu.cn, liuyumeng-2021@emails.bjut.edu.cn, \{yuanming\_li, hongji\_zeng\}@foxitsoftware.com}
}

%
%
%
\maketitle              

\begin{abstract}
With the rapid advancement of natural language processing (NLP) technologies, the demand for high-quality Chinese document question-answering datasets is steadily growing. To address this issue, we present the Chinese Multi-Document Question Answering Dataset(ChiMDQA), specifically designed for downstream business scenarios across prevalent domains including academic, education, finance, law, medical treatment, and news. ChiMDQA encompasses long-form documents from six distinct fields, consisting of 6,068 rigorously curated, high-quality question-answer (QA) pairs further classified into ten fine-grained categories. Through meticulous document screening and a systematic question-design methodology, the dataset guarantees both diversity and high quality, rendering it applicable to various NLP tasks such as document comprehension, knowledge extraction, and intelligent QA systems. Additionally, this paper offers a comprehensive overview of the dataset's design objectives, construction methodologies, and fine-grained evaluation system, supplying a substantial foundation for future research and practical applications in Chinese QA. The code and data are available at: \url{https://anonymous.4open.science/r/Foxit-CHiMDQA/}.

\end{abstract}
\section{Introduction}
The rapid development of NLP technology has significantly accelerated the application of intelligent question-answering (QA) systems. Breakthroughs in pre-trained language models, such as BERT \cite{wang2024utilizing} and the GPT\cite{rathje2024gpt} series, have unlocked remarkable potential in domains including information retrieval, intelligent customer service, and knowledge management. However, current research predominantly focuses on English-language scenarios, and the development of Chinese question-answering systems is confronted with numerous challenges.

Although notable datasets such as CMRC \cite{cui-etal-2019-span} and DuReader \cite{DuReader} have contributed to the Chinese QA field, they reveal notable deficiencies in document diversity and question type comprehensiveness. These inherent limitations restrict their effectiveness in navigating the complex demands of real-world, multi-domain, and multi-task environments. Consequently, the development of a robust, multi-dimensional Chinese QA dataset represents a pivotal advancement for expanding the frontiers of Chinese NLP research.

Early QA datasets exhibited significant limitations in document coverage and question design depth. Mainstream datasets like SQuAD\cite{rajpurkar-etal-2016-squad} predominantly rely on Wikipedia as a singular data source, resulting in a critical absence of specialized domain documents such as academic papers and legal texts. MS MARCO\cite{MS-MARCO}, while integrating web page data, presents substantial challenges with 90\% of documents in HTML format, impeding effective handling of nested tables and cross-page formulas in PDF documents. DuReader\cite{DuReader} introduces a real Chinese search corpus, yet its documents average less than 400 words and fail to encompass structured long-form texts like financial and medical reports. In the domain of long document processing, MMLongBench-Doc\cite{MMLONGBENCH-DOC} supports cross-page reasoning for 47.5 pages of English documents, but suffers from vague problem classification and limited cross-domain applicability. These fundamental constraints significantly hinder existing models' capabilities to parse multi-domain, rich-format long documents in real-world scenarios.

This paper presents the following contributions:
(1) Chinese Multi-Topic Long Document Coverage: A comprehensive integration of six thematic document types comprising academic, financial, legal, medical, educational, and news, containing 6,068 QA pairs, providing a high-quality, diverse data resource for Chinese document question-answering.
(2) Hierarchical Question Classification System: Grounded in the explicit and implicit fact theoretical framework\cite{rag}, the question system encompasses two primary categories and ten subcategories, achieving two significant breakthroughs: Expanding the explicit fact category beyond SQuAD\cite{rajpurkar-etal-2016-squad} and HotpotQA\cite{HotpotQA} by introducing three novel complex task types: filtering, statistical analysis, and computational reasoning; Developing implicit fact-based questions covering open-domain tasks such as generation and recommendation, extending assessment dimensions by 300\% compared to DuReader\cite{DuReader}'s fact-based approach.
(3) Fine-Grained Evaluation System: An innovative evaluation framework combining factual and open questions, comprising 21 comprehensive metrics for the Retrieval-Augmented Generation (RAG) system. This approach enables holistic performance assessment of QA systems across diverse question types, establishing a crucial foundation for Chinese QA research and application.

\begin{table}[!t]
\centering
\small
\caption{Comparison of ChiMDQA with existing QA datasets.}
\resizebox{\textwidth}{!}{
\begin{tabular}{lccccccc}
\toprule
\textbf{Datasets} & \textbf{Data Source} & \textbf{Size} & \textbf{Language} & \textbf{PDF-Based} & \textbf{Stratification} \\
\midrule
SQuAD\cite{rajpurkar-etal-2016-squad} & Wikipedia & 100,000 & English & No & No \\
MMLU-Pro\cite{MMLU-Pro} & STEM Website & 12,032 & English & No & No \\
Natural Questions\cite{natural-qa} & Google search, Wikipedia & 307,373 & English & No & No \\
MS MARCO\cite{MS-MARCO} & Bing’s search & 1,010,916 & English & No & No \\
TriviaQA\cite{joshi-etal-2017-triviaqa} & Wikipedia & 95,956 & English & No & No \\
HotpotQA\cite{HotpotQA} & Wikipedia & 113,000 & English & No & No \\
ScienceQA\cite{ScienceQA} & Elementary and high school science curricula & 21,208 & English & No & No \\
MULongBench-Doc\cite{MMLONGBENCH-DOC} & lengthy PDF-formatted documents & 1,091 & English & Yes & No \\
DuReader\cite{DuReader} & Baidu Search, Baidu Zhidao documents & 200,000 & Chinese & No & 6 \\
DRCD\cite{DRCD} & Wikipedia & 30,000 & Chinese & No & 7 \\
\midrule
\textbf{ChiMDQA} & lengthy PDF-formatted documents & 6,068 & Chinese & Yes & 10 \\
\bottomrule
\end{tabular}
}
\label{tab: related_work}
\end{table}

\section{ChiMDQA Dataset}
\subsection{Motivation}
As illustrated in Table \ref{tab: related_work}, existing QA datasets primarily focus single-topic documents, thereby significantly constraining their applicability to the complex, multi-domain document processing requirements of real-world scenarios. In addition, many of these datasets lack a fine-grained categorization of question types, failing to capture the full spectrum of question complexity—from straightforward fact retrieval to multi-step reasoning and semantic inference. To address these limitations, we propose ChiMDQA, a dataset designed to bridge this gap by incorporating multi-topic documents and a diverse set of well-defined, fine-grained question types. ChiMDQA is intended to serve as a more challenging and practically relevant benchmark, supporting the advancement of research and development in Chinese document-level QA.

\subsection{Document Topics and Question Types}

\subsubsection{Document Topics}
\label{topic}
The ChiMDQA dataset covers six representative domains, namely academic, education, finance, law, medical treatment, and news to ensure broad topical coverage and real-world applicability. Academic documents consist of peer-reviewed research papers, which disseminate research findings and support academic endeavors. Educational documents focus on textbooks and instructional materials, covering various disciplines to facilitate teaching and educational research. Financial documents include financial reports that reflect corporate economic activities, aiding in financial analysis and decision-making. Legal documents comprise legal documents essential for legal practice, research, and case management. Medical documents involve clinical guidelines that inform medical practice. News documents feature journalistic articles that disseminate societal information. These domains were selected for their topical richness, representativeness, and relevance to real-world applications. Collectively, they offer comprehensive coverage of core knowledge required across a wide range of QA scenarios.

\subsubsection{Question Types}
ChiMDQA's question typology is grounded in Microsoft's two-level framework of explicit and implicit facts \cite{rag}, refined for single-document QA. We use a two-tier system: Level 1 (L1) for explicit facts requiring direct extraction, and Level 2 (L2) for implicit facts needing inference or integration of information. For instance, "Where will the 2024 Olympics be held?" is an L1 question, while "What is the ruling party in the country where Canberra is located?" is an L2 question, as it requires synthesizing multiple pieces of information.

\begin{figure}[!t]
    \centering
    \resizebox{1\textwidth}{!}{
    \includegraphics[width=1\textwidth]{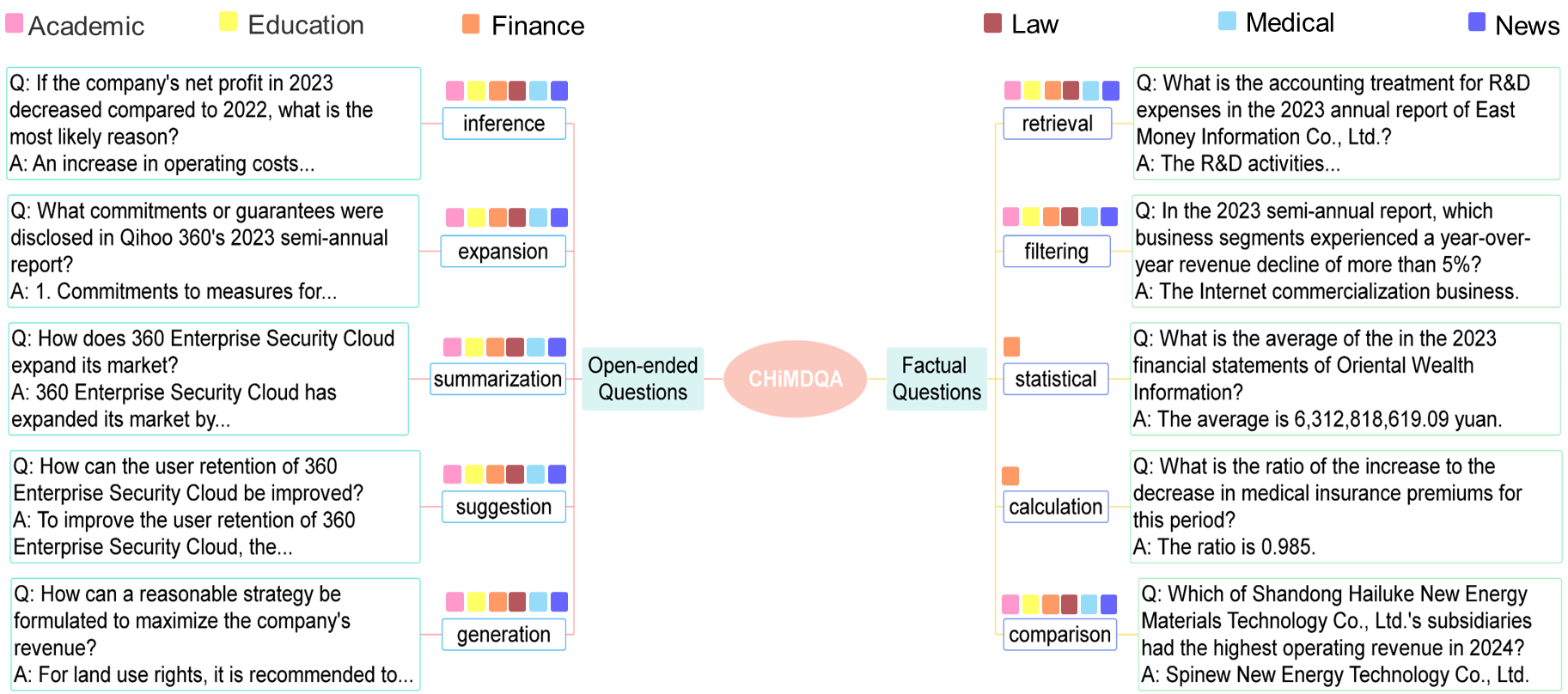}
    }
    \caption{Examples of Question Types and Their Topic Coverage in the ChiMDQA Dataset.}
    \label{QAexample}
 \end{figure}


QA pairs in ChiMDQA are evenly divided between L1 and L2 and further categorized into factual and open-ended types, refined into ten subtypes to align with downstream tasks. Examples are shown in Figure \ref{QAexample}.

\subsubsection{Factual Questions}
Factual questions are those whose answers can be directly extracted from the document or obtained through simple calculations or filtering, without subjective reasoning or generative capabilities. This category includes the following subtypes:\textbf{(1) Retrieval Questions:} These are straightforward queries requiring the extraction of specific information explicitly stated in the document. No inference or summarization is involved. \textbf{(2) Filtering Questions:} These involve selecting target entities based on multiple conditions, such as Boolean logic, numeric ranges, classification constraints, or relational attributes. \textbf{(3) Statistical Questions:} These require basic statistical analysis of document content, including operations such as computing frequency, mean, maximum, minimum, median, mode, range, variance, standard deviation, and coefficient of variation. \textbf{(4) Computational Questions:} These involve arithmetic manipulation of numerical data within the document, including addition, subtraction, multiplication, division, percentage, and ratio calculations. \textbf{(5) Comparison Questions:} These questions require comparing entities of the same type along one or more dimensions, often necessitating a lightweight evaluation framework to assess differences in characteristics.

\subsubsection{Open-ended Question}
Open-ended questions require reasoning, creative generation, or the integration of domain-specific knowledge. Answers are not constrained to a fixed template but instead emphasize semantic relevance, logical soundness, and contextual alignment. These questions typically demand a deeper understanding of document content and more comprehensive analytical capabilities. The open-ended category includes the following subtypes: \textbf{(1) Inference Questions:} These questions involve synthesizing dispersed information through logical reasoning and commonsense inference to uncover implicit facts. Examples include causal inference, multi-hop reasoning, or latent fact identification. \textbf{(2) Expansion Questions:} These require systematic elaboration of information to build hierarchical or structured representations, such as reconstructing organizational charts, taxonomies, or document section relationships. \textbf{(3) Summarization Questions:} These call for compressing and restructuring document content to distill key features or underlying patterns. Tasks may involve summarization, pattern identification, or concept clustering.\textbf{(4) Suggestion Questions:} These involve generating recommendations or decisions grounded in factual evidence from the document, often enhanced by domain-specific insights or contextual interpretation.\textbf{(5) Generation Questions:} These involve creative construction of new content based on the source document. Examples include imaginative extrapolation, cross-format transformation, or information synthesis.

In summary, ChiMDQA offers a comprehensive collection of documents across six domains, systematically categorizing questions into factual and open-ended types. By providing a structured approach to question answering, it provides a robust framework for evaluating and advancing question-answering methodologies across diverse computational linguistic research and applied scenarios.

\subsection{Dataset Construction}
\label{construct}
\begin{figure}[!t]
\centering
\resizebox{1\textwidth}{!}{
\includegraphics[width=1\textwidth]{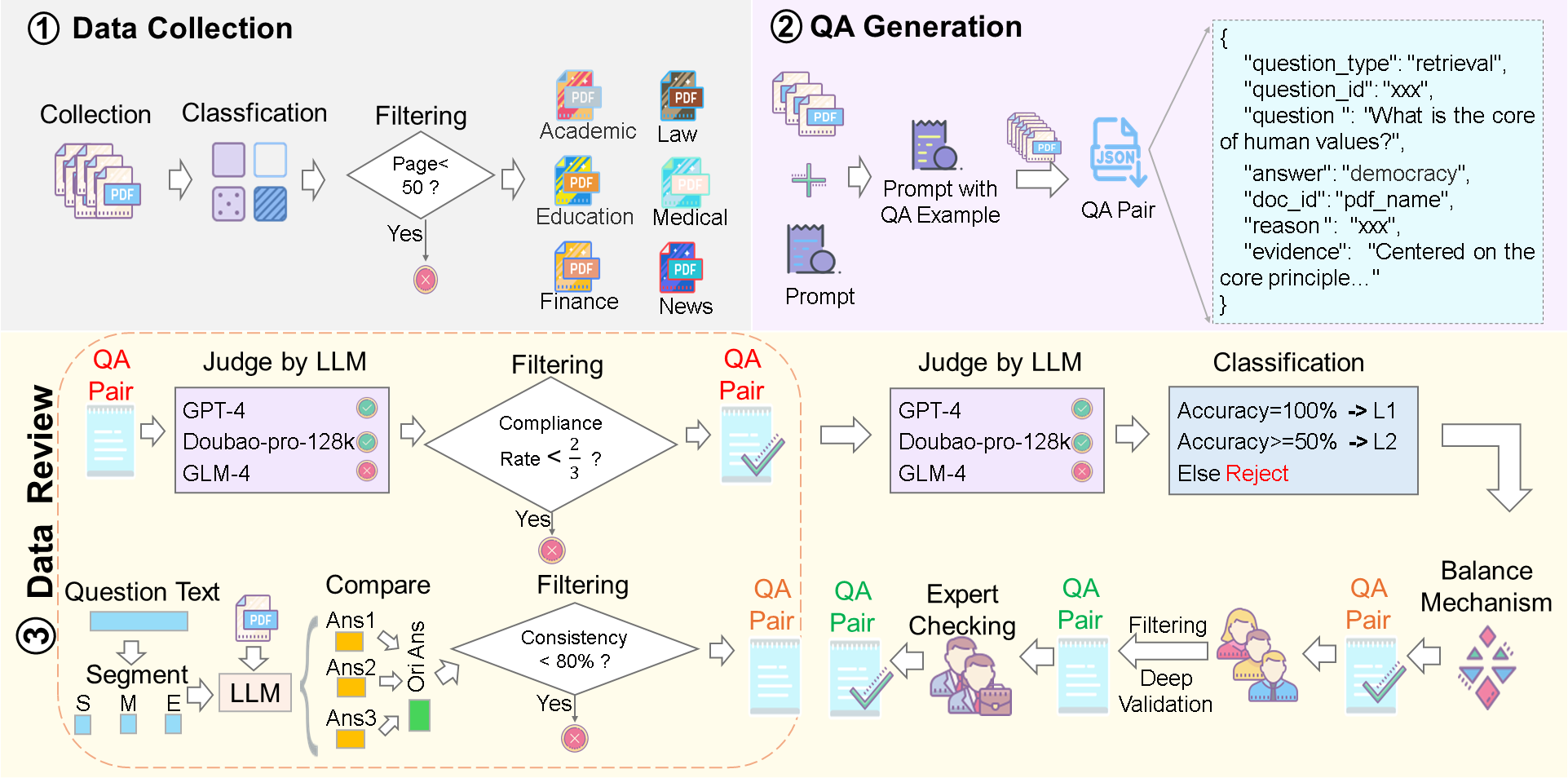}
}
\caption{The Framework of ChiMDQA Dataset Construction Process.(S:Start;M:Middle;E:End;Ori Ans:Original Answer)}
\label{framwork}
\end{figure}

ChiMDQA was constructed through a multi-stage pipeline, encompassing: data collection, QA pair generation, data review, and validation and statistics. An overview of the process is shown in Figure \ref{framwork}.

\subsubsection{Step 1: Data Collection} 
We collected approximately 15,000 multilingual PDF documents through web crawling and manual acquisition. Non-Chinese documents were filtered out using automated language detection tool. The remaining documents were further screened based on the following guidelines:\textbf{(1) File Format:} Only high-resolution, original (non-scanned) PDF files were retained to ensure extractable text and structural integrity. \textbf{(2) Timeliness and Relevance:} Documents were required to be published within the last five years and demonstrate clear relevance to one of the six target domains.\textbf{(3)Source Authority and Copyright Compliance:} All documents were obtained from credible sources with verified copyright compliance.

Following these criteria, we curated 60 representative documents—10 from each domain. Each selected PDF was processed through text extraction (using the PyMuPDF library), format normalization, and content cleaning. Text was segmented into logical blocks, and non-informative elements were removed via rule-based filtering to enhance data usability.

\subsubsection{Step 2: QA Pair Generation}
For each of the ten predefined question type, we designed specialized prompts that included task descriptions, generation requirements, JSON formatting specifications, and illustrative examples to guide large language models (LLMs) in producing high-quality QA pairs. After iterative refinement through pilot runs, we incorporated three validated JSON samples into the final prompt template used for large-scale generation.

We also conducted detailed document-level statistics, analyzing page count, word count, and token length across domains. Based on this analysis, we determined that models must support context windows of at least 64k tokens to effectively process long-form documents. As ChiMDQA is a Chinese-language dataset, we prioritized LLMs that have demonstrated strong performance in Chinese NLP and complex reasoning tasks. The selected models include: moonshot-v1-128k\cite{kimi}, doubao-pro-128k\cite{doubao}, qwen-plus\cite{qwen2023}, deepseek-chat\cite{deepseek} and glm-4-pro\cite{glm}. All models support context lengths of 64k tokens or more, making them well-suited for long-document comprehension.

Using the finalized prompts, we generated an initial batch of 250 QA pairs. We then randomly sampled 20\% of the generated pairs for manual evaluation to assess their alignment with quality standards. Based on performance, consistency, and overall generation quality, GLM-4-Pro\cite{glm} was selected as the final model for generating the full set of QA pairs.

\subsubsection{Step 3: Data Review}
To ensure the quality and reliability of ChiMDQA, we implemented a hybrid verification pipeline combining automated evaluation with human-in-the-loop review. This multi-layered framework was designed to rigorously assess the factual accuracy, coherence, and structural diversity of the generated QA pairs.

\textbf{Step 3.1: Automated Evaluation}
To support large-scale quality assurance and efficient filtering, we design a multi-dimensional collaborative auditing framework that balances both efficiency and accuracy. The system leverages heterogeneous model verification, context robustness testing, and a dynamic difficulty-balancing mechanism. The pipeline includes: \textbf{(1) Multi-model Collaborative Pre-screening:} Three heterogeneous LLMs—GPT-4\cite{gpt4}, Doubao-Pro-128k\cite{doubao}, and GLM-4\cite{glm}—are employed for parallel inference and validation of QA pairs. Each model assigns a confidence score to the generated answer. Only QA pairs that receive confidence scores above 0.85 from at least two models are retained for further review.
\textbf{(2) Context Sensitivity Screening:}To test robustness, we apply controlled context truncation (e.g., extracting the beginning, middle, or end of the document) and compare the model-generated responses across these segments to those generated from full context. QA pairs are flagged if answer consistency drops below an 80\% threshold.\textbf{(3)Difficulty Calibration and Distribution Balancing:} We construct a gradient-based test set that classifies questions into two predefined difficulty levels, where L1 encompasses explicit fact retrieval and L2 involves single-step reasoning or knowledge integration. The distribution of these levels was continuously monitored. If a significant imbalance was detected (e.g., L1 exceeding 60\%), sampling and generation strategies were adjusted accordingly to maintain diversity and prevent skewed learning dynamics.

\begin{table}[!t]
\centering
\small
\caption{Dataset statistics of ChiMDQA.}
\setlength{\tabcolsep}{8pt} 
\rowcolors{2}{gray!25}{white} 
\begin{tabular}{lp{4cm}|lp{4cm}}
\toprule
\textbf{Statistics} & \textbf{Number} & \textbf{Statistics} & \textbf{Number} \\
\midrule
\textbf{\#Problems} & 6,068 & \textbf{Length} & \\
\textbf{Topics} & & \textbf{Question Length} & \\
- Academic  & 922 & - \textit{maximum length} & 393 \\
- Education  & 1,028 & - \textit{minimum length} & 8 \\
- Finance & 980 & - \textit{avg length} & 37.31 \\
- Law & 972 & \textbf{Reference Answer Length} & \\
- Medical  & 1,107 & - \textit{maximum length} & 4,726 \\
\hspace{0.5em} treatment\newline &  & - \textit{minimum length} & 2 \\
- News & 1,059 & - \textit{avg length} & 162.44 \\
\bottomrule
\end{tabular}
\label{tab: detail_data1}
\end{table}

\begin{table}[!t]
\centering
\caption{Statistical Analysis of Document Details within the Dataset.}
\begin{tabularx}{\columnwidth}{l*{6}{>{\centering\arraybackslash}X}}
\toprule
\textbf{Topic} & \textbf{Max Pages} & \textbf{Min Pages} & \textbf{Avg Pages} & \textbf{Max Length} & \textbf{Min Length} & \textbf{Avg Length} \\
\midrule
\rowcolor{gray!25} 
Academic & 283 & 176 & 229.2 & 635,756 & 201,912 & 326,905 \\
Education & 129 & 60 & 102.1 & 154,363 & 65,796 & 97,356 \\
\rowcolor{gray!25} 
Finance & 275 & 118 & 167.7 & 247,067 & 107,265 & 156,131 \\
Law & 354 & 46 & 150.1 & 216,258 & 19,282 & 88,536 \\
\rowcolor{gray!25} 
Medical treatment& 181 & 64 & 102.5 & 335,273 & 4,061 & 96,088 \\
News & 160 & 60 & 100.7 & 728,287 & 317,266 & 593,027 \\
\rowcolor{gray!25} 
\midrule
Average & 230.33 & 87.33 & 142.13 & 386,167.33 & 119,263.67 & 226,340.43 \\
\bottomrule
\end{tabularx}
\label{tab: detail_data2}
\end{table}

\textbf{Step 3.2: Manual Review}
To further ensure the accuracy, clarity, and diversity of QA pairs, we conducted a comprehensive manual review following a strict five-stage cross-validation approach. The review procedure adhered to strict evaluation guidelines—namely, question clarity, answer correctness, and coverage across question types—and was implemented through a five-stage cross-validation framework:\textbf{(1) Initial Screening:}A team of three reviewers performed an initial evaluation using standardized templates to assess question clarity, grammatical correctness, and logical coherence. QA pairs that failed to meet baseline quality criteria were discarded.\textbf{(2) Deep Verification:}Two auditors independently examined the semantic accuracy of the answers, checked for logical coherence, and flagged ambiguous or unsupported claims. The focus was on ensuring alignment with the document and factual correctness. \textbf{(3) Dispute Arbitration:}In cases where reviewers disagreed, a third reviewer conducted a double-blind cross-validation, making the final decision on whether to retain, revise, or discard the QA pair. \textbf{(4) Diversity Review:}Reviewers assessed the semantic similarity of QA pairs to identify and flag redundant or duplicate questions. Based on their evaluations and optimization suggestions, the question generation strategy was refined to enhance diversity and reduce content overlap. \textbf{(5) Experts Validation:} A final round of domain-specific review was conducted by experts, who assessed the appropriateness and credibility of both questions and answers. They also resolved edge cases and resolved remaining disputes to ensure domain fidelity and factual accuracy.

\subsubsection{Step 4: Data Validation and Statistics}
To assess the quality and correctness of the dataset, we randomly sampled 100 QA pairs from each domain and conducted a manual validation process following the internal verification guideline and the Google Proof standard. The evaluation was carried out by members of the author team. Results indicated an overall error rate of approximately 3\%, with most questions requiring contextual understanding rather than relying on memorized or easily retrievable internet knowledge. In addition, a comprehensive review was performed to ensure that the dataset contained no sensitive personal information or copyright-infringing content.

Overall, statistics related to question length and reference answer length are summarized in Table~\ref{tab: detail_data1}, while document-level statistics are presented in Table~\ref{tab: detail_data2}.

\subsection{Future Expansion of the Dataset}
While the current version of ChiMDQA provides a robust foundation for evaluation, we recognize the importance of continuous growth to keep pace with the evolving capabilities of LLMs. We have established a clear roadmap for future expansion, focusing on both increasing the dataset's scale and broadening its scope. Our expansion strategy will follow a semi-automated pipeline: (1) \textbf{Domain Expansion:} We plan to incorporate new high-value domains such as engineering, environmental science, and government reports. (2) \textbf{Automated Candidate Generation:} We will leverage state-of-the-art LLMs to generate a large pool of candidate QA pairs for new documents, using the same structured prompting techniques developed for the initial dataset. (3) \textbf{Rigorous Human-in-the-Loop Verification:} Crucially, all automatically generated candidates will undergo the same multi-stage human review and validation process detailed in Section \ref{construct}. This ensures that any new additions meet the high standards of quality, accuracy, and diversity that define ChiMDQA. This hybrid approach will allow us to scale the dataset efficiently while maintaining its integrity as a reliable and challenging benchmark.

\section{Experiments}
\subsection{Experimental Setup}
We evaluate eight closed-source LLMs, each supporting a context window size of at least 128k tokens. The models include GPT-4\cite{gpt4}, GPT-4o\cite{gpt4o}, GLM-4-Plus\cite{glm}, GLM-4-Air\cite{glm},GLM-4-Flash\cite{glm}, YAYI-30B\cite{yayi}, Qwen-Plus\cite{qwen2023} and Doubao-Pro-128k\cite{doubao}. These models represent some of the most advanced LLMs currently available, primarily developed by the Chinese research and open-source communities.
In addition to these models, we also conducted preliminary tests on other leading models from large technology companies, such as Google's Gemini series \cite{gemini2024}, Baichuan's models \cite{baichuan2023}, and Meta's Llama series \cite{llama3-2024}. Our observations indicate that their performance trends are largely consistent with the results presented in this paper, with top-tier models like Gemini-1.5-Pro showing competitive performance comparable to GPT-4o. To maintain the clarity and focus of our result tables, we have centered our detailed analysis on the selected eight models, which provide a representative and diverse sample of the current state-of-the-art for Chinese long-document QA.
They provide a strong foundation for evaluating both non-RAG and RAG-based question-answering systems.

\subsection{Evaluation Metrics}
To comprehensively assess the performance of non-RAG and RAG systems, we introduce a suite of evaluation metrics tailored to each setting. These fine-grained metrics enable more precise measurement of model capabilities and serve as guidance for downstream optimization.

For Non-RAG Evaluation for Factual Questions, We adopt the following evaluation metrics: Correct (CO) \cite{simpleqa}: The predicted answer fully includes the reference answer and introduces no contradictory information. Not Attempted (NA)\cite{simpleqa}: The predicted answer does not include the reference answer but also does not introduce contradictions. Incorrect (IN)\cite{simpleqa}: The predicted answer contradicts the reference answer. Correct Given Attempted (CGA)\cite{simpleqa}: The proportion of correctly answered questions among all attempted questions. F1-Score: The harmonic mean between Correct and CGA, calculated as:\\
\begin{equation}
\mathrm{F1-Score}=2\times\frac{\mathrm{CO}\times \mathrm{CGA}}{\mathrm{CO}+\mathrm{CGA}}
\end{equation}

For open-ended questions, we apply the following automatic evaluation metrics: METEOR (M) \cite{banerjee-lavie-2005-meteor}: Measures semantic similarity between generated and reference answers. ROUGE-L (R-L) \cite{banerjee-lavie-2005-meteor}: Evaluates lexical overlap based on the longest common subsequence. CIDEr (C) \cite{CIDEr}: Captures consensus and semantic consistency between generated and reference answers. Perplexity (PPL) \cite{brown-etal-1992-class}: Reflects the model's confidence when generating text; lower values indicate more fluent and confident output. BERTScore-F1 (B-F1) \cite{BERTScore}: Assesses semantic similarity using contextual embeddings from BERT. We use the bert-base-chinese model in this study.
The BERTScore-F1 is calculated using the following formulas:\\
\begin{flalign}
&\hspace{4.8cm}\text{B-Precision} = \frac{\sum_{i} \operatorname{sim}\left(g_{i}, r_{i}\right) \times \mathbb{I}\left(g_{i} \in r\right)}{\sum_{i} \mathbb{I}\left(g_{i} \in g\right)} &\label{eq:precision} \\
&\hspace{5cm}\text{B-Recall} = \frac{\sum_{i} \operatorname{sim}\left(g_{i}, r_{i}\right) \times \mathbb{I}\left(r_{i} \in r\right)}{\sum_{i} \mathbb{I}\left(r_{i} \in r\right)} &\label{eq:recall} \\
&\hspace{5cm}\text{B-F1} = 2 \times \frac{\text{B-Precision} \times \text{B-Recall}}{\text{B-Precision} + \text{B-Recall}} &\label{eq:f1}
\end{flalign}

Here, $g_{i}$and $r_{i}$ represent tokens in the generated and reference answers, respectively. $\operatorname{sim}\left(g_{i}, r_{i}\right)$ denotes the cosine similarity between token embeddings, and $\mathbb{I}(\cdot)$  is the indicator function.

For RAG systems, in addition to adopting the same evaluation metrics as non-RAG systems for factual and open-ended questions, we also incorporate the RAGChecker\cite{RAGchecker} framework to conduct fine-grained evaluation of the retrieval and generation modules. The retrieval module metrics\cite{RAGchecker} include “Claim Recall” and “Context Precision”, which measure the coverage and precision of the retrieved information required to generate a correct answer. The generation module metrics\cite{RAGchecker} include Faithfulness, which assesses the extent to which the generated answer relies on the retrieved content; Relevant Noise Sensitivity and Irrelevant Noise Sensitivity, which evaluate the model's sensitivity to semantically related and unrelated distractors, respectively; Hallucination, which detects statements in the generated response that are unrelated to both the reference answer and the retrieved context; Self-Knowledge, which measures the model's ability to answer based on its own pretrained knowledge; and Context Utilization, which evaluates how effectively the retrieved content is used during generation. Overall metrics\cite{RAGchecker} are computed by calculating the proportion of correct factual claims in the predicted answer (Precision) and the proportion of reference claims that are correctly predicted (Recall), and further integrating them into a comprehensive F1-Score to assess the quality of the generated response. 

In summary, our evaluation framework offers a comprehensive, multi-perspective approach to measuring both the retrieval and generation performance of QA systems, supporting rigorous assessment and optimization.

\section{Results and Analysis}
\subsection{Baselines}
\begin{table}[!t]
\centering
\fontsize{10}{10}\selectfont
\caption{Results of different models on factual questions in the ChiMDQA dataset. For metrics, \textbf{CO}, \textbf{NA}, \textbf{IN}, \textbf{CGA}, and \textbf{F1} denote ``Correct'', ``Not attempted'', ``Incorrect'', ``Correct given attempted'', and ``F1-Score'', respectively. \textbf{Bold} denotes the best model scores, and \underline{underline} denotes the second-best model scores.}
\begin{tabular}{l|*{5}{w{c}{1cm}}|*{6}{w{c}{1.2cm}}}
\toprule
\multirow{2}{*}{\textbf{Models}} & \multicolumn{5}{c|}{\textbf{Overall results on 5 metrics}}  & \multicolumn{6}{c}{\textbf{F1-Score$^\uparrow$ on 6 topics}}\\
\cmidrule(lr){2-12} 
 &\textbf{CO}$^\uparrow$     & \textbf{NA}$^\downarrow$    & \textbf{IN}$^\downarrow$     & \textbf{CGA}$^\uparrow$  & \textbf{F1}$^\uparrow$     & \textbf{Acad}        & \textbf{Edu}  & \textbf{Fin}     & \textbf{Law}        & \textbf{Med} &\textbf{News}   \\ 
\midrule
\textbf{GPT-4\cite{gpt4}} & 66.7 & 11.3 & 22.0 & 74.2 & 70.5 & 63.5 & 87.8 & 53.9 & 84.5 & 67.0 & 73.0 \\

\textbf{GPT-4o\cite{gpt4o}} & \textbf{73.2} & \textbf{9.3} & 17.5 & 81.4 & \textbf{76.5} & 69.0 & 90.7 & \textbf{57.6} & \textbf{87.4} & 74.9 & \textbf{77.8} \\

\textbf{GLM-4-Plus\cite{glm}} & \underline{69.5} & \underline{10.0} & 20.5 & 77.0 & 73.2 & 65.8 & 89.2 & 56.0 & \underline{86.0} & 70.0 & \underline{74.5} \\

\textbf{GLM-4-Air\cite{glm}} & 67.1 & 10.2 & 22.8 & 74.7 & 70.6 & 64.0 & 87.2 & 54.0 & 85.3 & 67.5 & 73.5 \\

\textbf{GLM-4-Flash\cite{glm}} & 61.8 & 15.4 & 22.8 & 73.1 & 65.9 & \textbf{71.2} & 89.5 & 46.5 & 80.5 & 70.9 & 50.7 \\

\textbf{YAYI-30B\cite{yayi}} & 61.4 & 24.8 & \underline{13.8} & \underline{81.6} & 70.1 & 44.8 & \underline{94.6} & \underline{56.3} & 81.2 & 77.8 & 64.5 \\

\textbf{Qwen-Plus\cite{qwen2023}} & 68.5 & 17.9 & \textbf{12.6} & \textbf{84.7} & \underline{76.3} & 61.5 & \textbf{97.4} & 52.1 & 84.9 & \textbf{87.5} & 60.6 \\

\textbf{Doubao-Pro-128k\cite{doubao}} & 69.4 & 11.4 & 19.2 & 78.3 & 73.6 & \underline{70.4} & 89.7 & 54.5 & 78.9 & \underline{79.5} & 65.7 \\
\bottomrule
\end{tabular}
\label{tab:factualmetrics}
\end{table}

\begin{table}[!t]
\centering
\fontsize{10}{10}\selectfont
\caption{Results of different models on open-ended questions in the ChiMDQA dataset. For metrics, \textbf{M}, \textbf{R-L}, \textbf{C}, \textbf{PPL}, and \textbf{B-F1} denote ``METEOR'', ``ROUGE-L'', ``CIDEr'', ``Perplexity'', and ``BERTScore-F1'', respectively. \textbf{Bold} denotes the best model scores, and \underline{underline} denotes the second-best model scores.}
\begin{tabular}{l|*{5}{w{c}{1cm}}|*{6}{w{c}{1.2cm}}}
\toprule
\multirow{2}{*}{\textbf{Models}} & \multicolumn{5}{c|}{\textbf{Overall results on 5 metrics}}  & \multicolumn{6}{c}{\textbf{BERTScore-F1$^\uparrow$ on 6 topics}}\\
\cmidrule(lr){2-12} 
 & \textbf{M}$^\uparrow$ & \textbf{R-L}$^\uparrow$ & \textbf{C}$^\uparrow$ & \textbf{PPL}$^\downarrow$  & \textbf{B-F1}$^\uparrow$ & \textbf{Acad}        & \textbf{Edu}  & \textbf{Fin}     & \textbf{Law}        & \textbf{Med} &\textbf{News}   \\ 
\midrule
\textbf{GPT-4\cite{gpt4}} & 24.9 & 29.8 & 34.7 & 30.3 & 74.9 & 74.8 & 77.9 & 76.8 & 77.6 & 76.5 & 75.3 \\
\textbf{GPT-4o\cite{gpt4o}} & \textbf{27.7} & \textbf{33.2} & \textbf{37.0} & 25.8 & \textbf{81.2} & \textbf{80.9} & \underline{81.9} & \textbf{79.7} & \textbf{81.3} & \textbf{79.8} & \textbf{80.9} \\
\textbf{GLM-4-Plus\cite{glm}} & \underline{27.4} & \underline{32.9} & \underline{36.9} & \underline{25.0} & \underline{80.4} & \underline{80.5} & 81.5 & \underline{79.4} & 80.7 & \underline{79.6} & \underline{80.7} \\ 
\textbf{GLM-4-Air\cite{glm}} & 25.4 & 31.9 & 27.7 & 30.1 & 78.5 & 73.0 & 81.3 & 78.5 & 80.5 & 78.9 & 78.6 \\
\textbf{GLM-4-Flash\cite{glm}} & 26.0 & 32.5 & 28.0 & 27.8 & 79.3 & 74.0 & \textbf{82.0} & 79.0 & \underline{81.0} & 79.5 & 79.5 \\
\textbf{YAYI-30B\cite{yayi}} & 26.3 & 25.7 & 14.4 & \textbf{20.8} & 78.0 & 75.4 & 79.6 & 77.9 & 80.0 & 76.8 & 78.9 \\
\textbf{Qwen-Plus\cite{qwen2023}} & 26.0 & 28.2 & 18.2 & \underline{25.0} & 78.0 & 73.9 & 80.8 & 77.7 & 79.9 & 77.2 & 79.0\\
\textbf{Doubao-Pro-128k\cite{doubao}} & 22.7 & 26.4 & 12.3 & 53.1 & 76.1 & 74.5 & 78.4 & 75.3 & 73.7 & 76.0 & 78.5 \\
\bottomrule
\end{tabular}
\label{tab:freemetrics}
\end{table}

Table \ref{tab:factualmetrics} and table \ref{tab:freemetrics} present the evaluation results of various LLMs on the ChiMDQA dataset. These tables provide overall performance metrics for both factual and open-ended questions, including five evaluation indicators, as well as F1-Score and BERTScore-F1 across six distinct domains. From these results, we observe several insightful and noteworthy findings.
\begin{itemize}
\item[] \textbullet \quad \textbf{GPT-4o Achieves Superior Overall Performance}:As shown in Table \ref{tab:factualmetrics}, GPT-4o achieves the highest scores in overall metrics for factual questions, with a CGA rate of 81.4 and an F1 of 76.5. These results indicate that GPT-4o not only attempts more questions correctly but also maintains high precision, outperforming other leading models in factual QA tasks. Table \ref{tab:freemetrics} further reveals that GPT-4o excels in open-ended questions, achieving top scores in M, R-L, C, and B-F1, with a B-F1 of 81.2. These results demonstrate GPT-4o's superior generative capabilities and semantic fluency.

\item[] \textbullet \quad \textbf{High Perplexity Across Models:}:Table \ref{tab:freemetrics} indicates that all models exhibit relatively high PPL scores on open-ended questions. For instance, GLM-4-Flash records a PPL of 27.8, while Doubao-Pro-128k reaches 53.1. High perplexity values suggest greater uncertainty in text generation, likely attributable to the inherent diversity of open-ended questions, which often lack a single correct answer. Consequently, models must generate responses within a broad semantic spectrum, leading to increased uncertainty.

\item[] \textbullet \quad \textbf{Variability in Domain-Specific Performance}:Model performance varies significantly across domains. For example, in the financial domain, YAYI-30B achieves a B-F1 of 77.9 on open-ended questions but only an F1 of 56.3 on factual questions. Conversely, Qwen-Plus attains an F-Score of 84.7 on factual questions in the legal domain but records a BERTScore of just 79.9 on open-ended questions. This variability underscores the limitations of current models in handling domain-specific knowledge and highlights the need for further optimization to enhance adaptability and accuracy in specialized fields.
\end{itemize}

\begin{table}[!t]
\centering
\fontsize{9}{9}\selectfont
\caption{Overall, Retrieval, and Generator Metrics for RAG Systems on ChiMDQA. For retrieval metrics, \textbf{Claim-R} and \textbf{Context-P} denote ``Claim Recall'' and ``Context Precision'', respectively. For generator metrics, \textbf{Fth}, \textbf{RNS}, \textbf{INS}, \textbf{Hlc}, \textbf{SK}, and \textbf{CU} denote ``Faithfulness'', ``Relevant Noise Sensitivity'', ``Irrelevant Noise Sensitivity'', ``Hallucination'', ``Self-knowledge'', and ``Context Utilization'', respectively. \textbf{Bold} denotes the best model scores, and \underline{underline} denotes the second-best model scores.}
\begin{tabular}{l|ccc|cc|cccccc}
\toprule
\multirow{2}{*}{\textbf{Models}} & \multicolumn{3}{c|}{\textbf{Overall Metrics}} & \multicolumn{2}{c|}{\textbf{Retrieval Metrics}} & \multicolumn{6}{c}{\textbf{Generator Metrics}} \\
\cmidrule{2-4} \cmidrule{5-6} \cmidrule{7-12}
 & \textbf{Precision}$^\uparrow$ & \textbf{Recall}$^\uparrow$ & \textbf{F1-Score}$^\uparrow$ & \textbf{Claim-R}$^\uparrow$ & \textbf{Context-P}$^\uparrow$ & \textbf{Fth}$^\uparrow$ & \textbf{RNS}$^\downarrow$ & \textbf{INS}$^\downarrow$ & \textbf{Hlc}$^\downarrow$ & \textbf{SK}$^\downarrow$ & \textbf{CU}$^\uparrow$ \\
\midrule
\textbf{GPT-4\cite{gpt4}} & 37.8 & 39.7 & 31.9 & 59.1 & 50.2 & 50.3 & 24.8 & 6.9 & 26.2 & 4.9 & 68.4 \\
\textbf{GPT-4o\cite{gpt4o}} & \textbf{43.9} & \underline{44.4} & \textbf{37.8} & \textbf{63.7} & \textbf{54.7} & \underline{57.6} & 25.9 & 7.8 & 22.3 & 5.3 & \textbf{74.6} \\
\textbf{GLM-4-Plus\cite{glm}}  & 43.0 & 43.5 & \underline{36.5} & \underline{62.3} & \underline{53.5} & 54.0 & 27.0 & 7.0 & \underline{21.0} & 6.0 & \underline{72.0} \\
\textbf{GLM-4-Air\cite{glm}} & 42.7 & 44.0 & 35.7 & 59.6 & 52.8 & 53.6 & 26.5 & 7.5 &\textbf{20.6} & 5.5 & 71.8 \\
\textbf{GLM-4-Flash\cite{glm}} & 43.1 & 38.9 & 31.3 & 59.4 & 50.6 & 46.5 & \textbf{22.0} & \underline{6.6} & 27.0 & 5.6 & 66.3 \\
\textbf{YAYI-30B\cite{yayi}} & 37.7 & \textbf{49.6} & 36.1 & 59.9 & 52.6 & \textbf{59.6} & 27.2 & 8.0 & 24.0 & 4.8 & 68.6 \\
\textbf{Qwen-Plus\cite{qwen2023}} & 40.6 & 43.1 & 34.1 & 59.2 &52.3 & 54.4 & 26.4 & 7.3 & 23.3 & \underline{4.4} & 71.0 \\
\textbf{Doubao-Pro-128k\cite{doubao}} & \underline{43.7} & 38.0 & 28.3 & 61.3 & 52.5 & 45.1 & \underline{23.9} & \textbf{6.2} & 23.3 & \textbf{4.2} & 70.0 \\
\bottomrule
\end{tabular}
\label{tab:ragmetrics}
\end{table}

\subsection{Retrieval-Augmented Baseline Model}
In this study, we explore the effectiveness of the RAG strategy in improving the accuracy of LLMs on different types of questions. Specifically, we implemented a RAG system based on Langchain\cite{langchain2023}, with the following improvements:
\begin{itemize}
    \item[] \textbullet \quad \textbf{Replacement of Embedding Model}: We replaced the original piccolo-base-zh model with the BCEmbedding model. This improvement aims to enhance the accuracy of vector representation and retrieval efficiency, thereby improving the model's ability to understand input queries.
    \item[] \textbullet \quad \textbf{Introduction of a Rerank Model}: We introduced the bce-reranker-base model to rerank the retrieved document snippets. This step helps to improve the relevance of the retrieval results, ensuring that the model can obtain information from the most relevant document snippets.
    \item[] \textbullet \quad \textbf{Implementation of Intelligent Segmentation}: We implemented intelligent segmentation features, including automatic completion of affiliations for headings, paragraph titles, and hierarchical levels. This feature helps the model to better understand and organize the document structure, thereby improving the accuracy of retrieval and generation.
\end{itemize}

Using the RAGChecker framework, we evaluate system performance (see Table \ref{tab:ragmetrics}). Overall, GPT-4o achieved the best scores across all metrics, with Precision, Recall, and F1-Score reaching 43.9, 44.4, and 37.8, respectively, indicating superior alignment between generated answers and factual content. However, none of the models surpassed 40 in F1-Score, suggesting that RAG systems still have room for improvement when handling complex questions. In terms of retriever performance, GPT-4o scored 63.7 and 54.7 in Claim Recall and Context Precision, respectively, while GLM-4-Plus closely followed, demonstrating strong accuracy in identifying relevant document segments. For generator performance, YAYI-30B achieved the highest Faithfulness score (59.6), reflecting strong reliability of generated answers, whereas GPT-4o scored highest in Context Utilization (74.6), indicating its ability to more effectively leverage retrieved content for answer generation. Nevertheless, all models exhibited a Hallucination rate exceeding 20, revealing that RAG systems still face challenges in controlling factuality and hallucinations in generated content.\\

\begin{figure}[!t]
    \centering
    \includegraphics[width=1\textwidth]{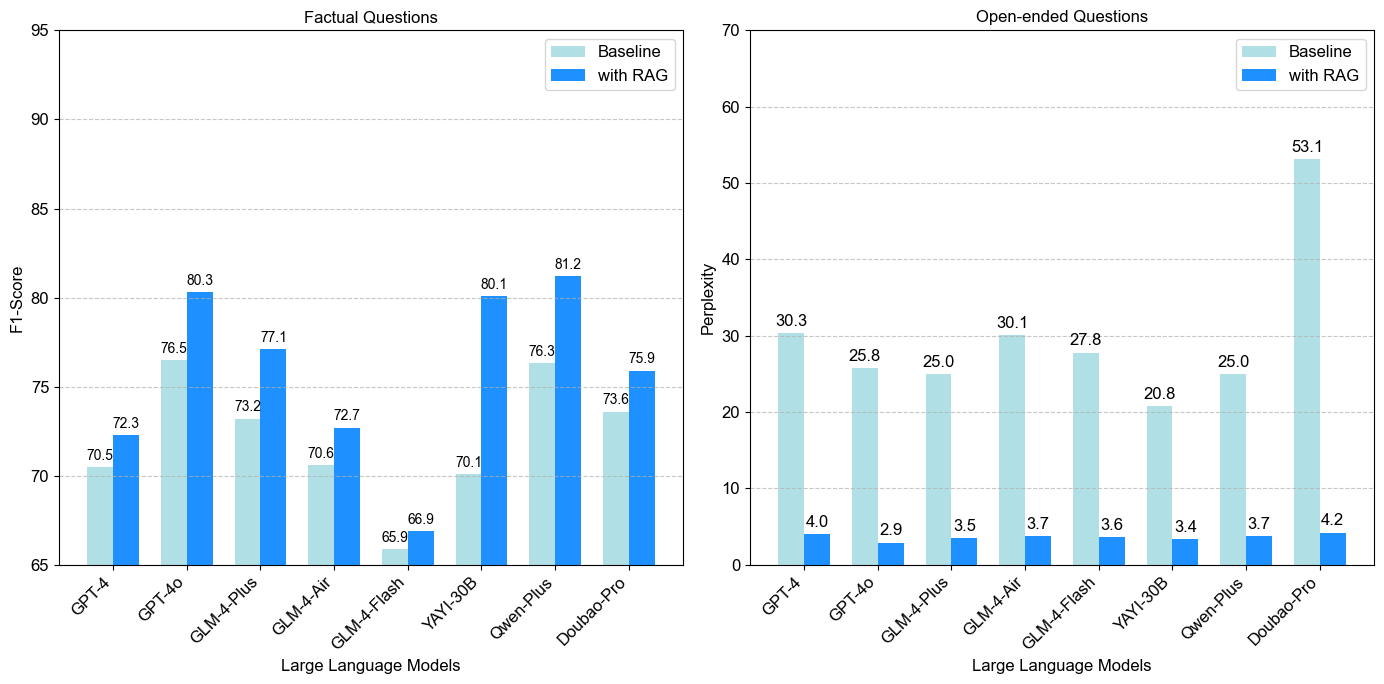}
    \caption{Performance Comparison of Models with and without RAG.}
    \label{fig:comparsion}
 \end{figure}

As shown in Figure \ref{fig:comparsion}, all models showed notable improvements under the RAG setting. Specifically, for factual questions, all models saw improvements in F1-Score after applying the RAG strategy. For instance, GPT-4o improved from 76.5 to 81.2, GLM-4-Plus from 73.2 to 77.1, and YAYI-30B from 70.1 to 74.0, with an average gain of 4.6\%. This demonstrates that RAG strategies effectively enhance model comprehension and factual answer generation. For open-ended questions, RAG significantly reduced model perplexity, indicating lower uncertainty and complexity in generated text. For example, Doubao-Pro-128k's perplexity dropped from 53.1 to 42.0, and GPT-4's from 30.3 to 24.0. Overall, the average perplexity decreased by 81.2\%, suggesting that RAG not only improves the accuracy of generated answers, but also reduces model uncertainty, thereby yielding more fluent and coherent outputs for both factual and open-ended question types.

\subsection{Domain-wise Results Analysis}
 \begin{figure}[!t]
    \centering
    \includegraphics[width=1\textwidth]{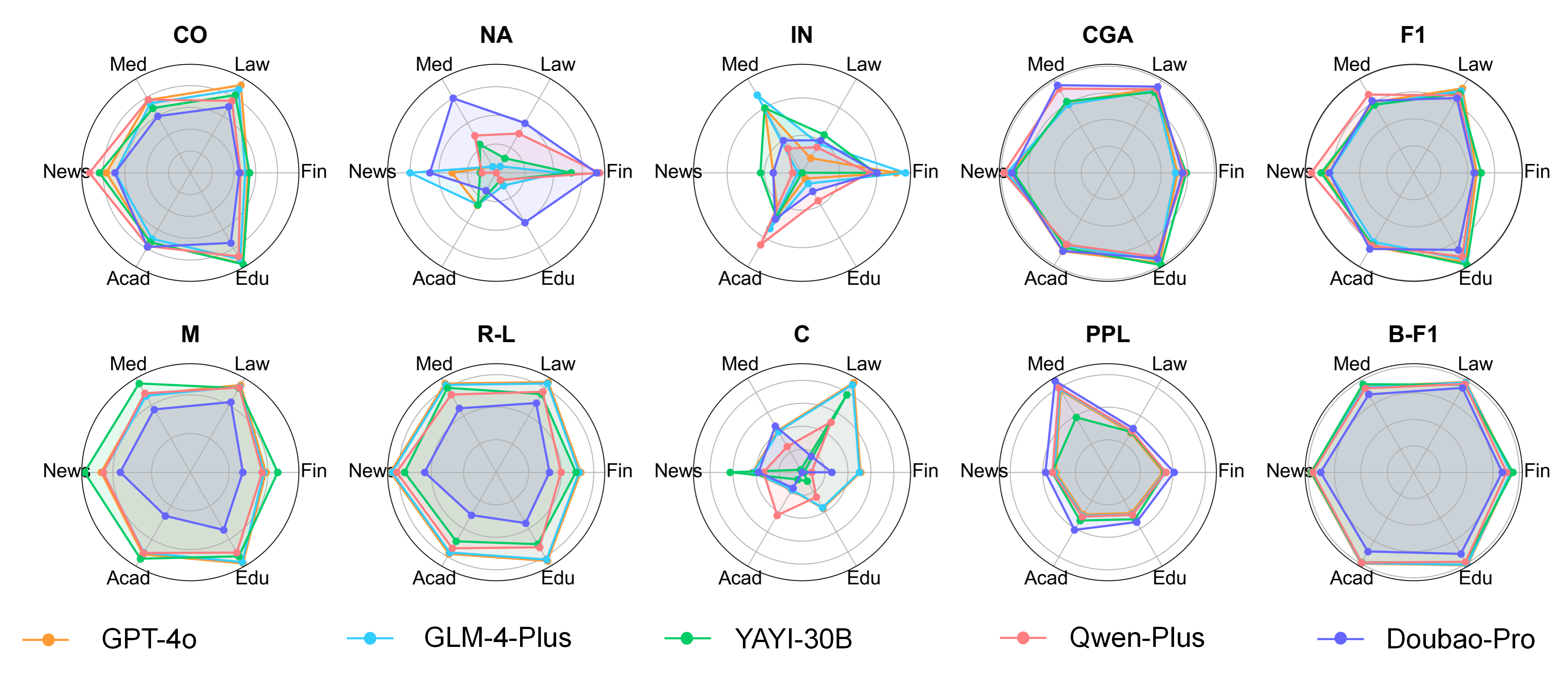}
    \caption{Results of different models for six topics. (Top: Evaluation metrics for factual questions;  Bottom: Evaluation metrics for open-ended questions)}
    \label{fig:topic}
 \end{figure}

As discussed in Section \ref{topic}, ChiMDQA spans six major domains, allowing for comprehensive assessment of model knowledge across diverse fields. Figure \ref{fig:topic} presents a comparison of GPT-based models across all domains under the RAG setting.

GPT-4o consistently outperformed other models across all domains, achieving high scores in both CO and F1 metrics. For instance, in the legal domain, GPT-4o achieved a CO of 92.84 and an F1 of 90.32. GLM-4-Plus also demonstrated competitive performance, especially in CGA and B-F1, showing high alignment between generated answers and both factual and contextual content. YAYI-30B and Qwen-Plus performed well in some domains, but had limited knowledge coverage in areas such as finance. Doubao-Pro-128k exhibited lower performance across domains, with a relatively high Incorrect rate and a higher proportion of NA answers, indicating poor knowledge retrieval and question interpretation in certain areas.

For open-ended questions, models performance varied more significantly across domains. GPT-4o and GLM-4-Plus achieved better performance in metrics like M and R-L, indicating their responses were more consistent and better aligned with reference answers. YAYI-30B and Qwen-Plus also performed reasonably well in some domains, though they exhibited more volatility in performance across tasks. Doubao-Pro-128k showed the weakest overall results in open-ended tasks, with low scores in M and R-L, indicating significant challenges in generating coherent and relevant long-form responses. In conclusion, GPT-4o and GLM-4-Plus demonstrated more balanced and robust performance across both question types and domains, whereas other models showed more variation in quality depending on the specific task or field.

\section{Conclusion}
The construction of the ChiMDQA dataset aims to address the limitations in document types and question diversity within existing Chinese QA datasets, thereby satisfying the complex multi-domain and multi-type QA requirements in real-world scenarios. Through rigorous document screening and a systematic question design process, ChiMDQA contains six fields and 6,068 high-quality QA pairs. The design, construction, and fine-grained evaluation system of the dataset lay a solid foundation for related research and applications in Chinese QA.
Experimental findings demonstrate that ChiMDQA effectively evaluate the processing capabilities of LLMs across different question types and business domains. Our exploration of the RAG strategy revealed that it substantially improves the models' F1-Score on factual questions and mitigates uncertainty for open-ended questions, thereby enhancing the accuracy and coherence of generated responses. However, there remains scope for improvement in handling complex questions and alleviating hallucinations.


\begin{credits}
\subsubsection{\ackname} 
This study was funded by Foxit Software Co. Ltd, Fuzhou, China. We are grateful to our colleagues at Foxit Software Co. Ltd, Fuzhou, China, for their valuable support and contributions. Special thanks also go to the reviewers for their insightful feedback.

\end{credits}

\fontsize{9}{9}\selectfont
\bibliographystyle{splncs04}
\bibliography{refs}

\end{document}

%% file: math_commands.tex
\usepackage{amsmath,amsfonts,bm}
\usepackage{multicol}

\def\eqref#1{equation~\ref{#1}}

\def\1{\bm{1}}

\DeclareMathAlphabet{\mathsfit}{\encodingdefault}{\sfdefault}{m}{sl}
\SetMathAlphabet{\mathsfit}{bold}{\encodingdefault}{\sfdefault}{bx}{n}

